\documentclass[preprint,12pt,lefttitle]{elsarticle}

\usepackage{amssymb}
\usepackage{amsmath}
\usepackage{graphicx}
\usepackage{booktabs}
\usepackage[hidelinks]{hyperref}

\usepackage{ragged2e}
\usepackage{etoolbox}
\makeatletter
\patchcmd{\pprintMaketitle}{\footnotesize\itshape\elsaddress}{\footnotesize\itshape\RaggedRight\elsaddress}{}{}
\patchcmd{\MaketitleBox}{\footnotesize\itshape\elsaddress}{\footnotesize\itshape\RaggedRight\elsaddress}{}{}
\def\ps@pprintTitle{%
  \let\@oddhead\@empty
  \let\@evenhead\@empty
  \def\@oddfoot{\centerline{\thepage}}%
  \let\@evenfoot\@oddfoot
}
\makeatother

\begin{document}

\begin{frontmatter}

\title{Backbone-Conditional Behavior of Modality Gating in Multi-Modal Prostate MRI Segmentation: A 5-Fold Cross-Validation and Gate Mechanism Analysis}

\author[inst1,inst3,inst4,inst5]{Yongbo Shu}
\author[inst2,inst3,inst4,inst5]{Wenzhao Xie}
\author[inst2,inst4,inst5]{Shanhu Yao}
\author[inst1,inst4,inst5]{Zirui Xin}
\author[inst1,inst4,inst5]{Luo Lei}
\author[inst3]{Kewen Chen}
\author[inst1,inst3,inst4,inst5]{Aijing Luo\corref{cor1}}
\ead{luoaj@csu.edu.cn}
\cortext[cor1]{Corresponding author}

\affiliation[inst1]{organization={The Second Xiangya Hospital of Central South University}, city={Changsha}, postcode={410011}, state={Hunan}, country={China}}
\affiliation[inst2]{organization={The Third Xiangya Hospital of Central South University}, city={Changsha}, postcode={410013}, state={Hunan}, country={China}}
\affiliation[inst3]{organization={School of Life Sciences, Central South University}, city={Changsha}, postcode={410013}, state={Hunan}, country={China}}
\affiliation[inst4]{organization={Hunan Provincial Key Laboratory of Medical Information Research (Central South University)}, city={Changsha}, postcode={410011}, state={Hunan}, country={China}}
\affiliation[inst5]{organization={Hunan Provincial Clinical Medical Research Center for Cardiovascular Intelligent Medicine}, city={Changsha}, postcode={410011}, state={Hunan}, country={China}}

\begin{abstract}
\textbf{Background and Objectives:} Robust segmentation of clinically significant prostate cancer (csPCa) on multi-parametric MRI must tolerate frequent degradation of its most informative diffusion sequences. Multi-modal fusion commonly employs learned modality gating under the assumption that gates implement per-sample modality quality routing. We challenge this assumption: it has rarely been empirically tested, and the conditions under which gating succeeds or fails across backbone architectures remain poorly characterized.

\textbf{Methods:} We conduct a systematic mechanism analysis of modality-isolated gated fusion (MIGF) for clinically significant prostate cancer (csPCa) segmentation across two backbone architectures (nnU-Net and Mamba) on the PI-CAI dataset ($n=1500$) with cross-cohort validation on Prostate158 ($n=158$). The protocol comprises (i) a factorial ablation over eight module configurations (gating, modality dropout, deep supervision) under 5-fold cross-validation across both backbones, totaling 180 trained models, and (ii) gate variance and counterfactual analysis on 30 trained gating models to characterize learned gate behavior.

\textbf{Results:} Modality gating exhibits backbone-conditional behavior. Adding gating significantly reduces nnU-Net's ranking score (gating--ablation configurations A2/A3/A4 versus baseline: $-0.036$/$-0.060$/$-0.054$, all $p<0.05$, paired $t$-test; marginal gating effect $-0.037$), whereas the G+M configuration significantly improves Mamba's ranking score ($+0.024$, $p=0.037$), with a directional but non-significant case-level specificity gain ($+0.114$, $p=0.067$). Gate-weight analysis reveals an $\approx$11$\times$ asymmetry in per-sample variability: nnU-Net gates collapse into a static modality prior (across-case SD $0.0033 \pm 0.0008$), while Mamba gates retain sample-dependent variation ($0.0365 \pm 0.0080$), with non-overlapping distributions. Counterfactual replacement with training-set means causes negligible change on nnU-Net ($\Delta$Score $+0.001$, n.s.) but consistent degradation on Mamba ($\Delta$Dice $+0.043$, 15/15). Modality dropout is the only component beneficial across both tested backbones ($+0.023$ and $+0.021$ marginal effect on ranking score). Cross-cohort evaluation reveals architecture-conditional specificity collapse: nnU-Net drops to case-level specificity $\approx 0$, whereas Mamba-based models retain non-zero specificity, with MIGF-Mamba highest among the tested configurations ($0.31$).

\textbf{Conclusions:} Learned modality gates do not universally implement per-sample quality routing; their effective behavior is conditional on the base architecture's inherent modality awareness. Among the tested configurations, MIGF-Mamba emerges as the most cross-cohort robust architecture for prostate MRI segmentation, while training-time modality dropout is the only component beneficial across both backbones.
\end{abstract}

\begin{highlights}
\item Learned modality gates behave conditionally on the backbone architecture.
\item On nnU-Net, gates collapse into a static modality prior (11x lower variability).
\item Counterfactual mean-gate replacement leaves nnU-Net performance unchanged.
\item MIGF-Mamba attained the highest case-level specificity under cross-cohort shift.
\item Modality dropout is the only component beneficial across both tested backbones.
\end{highlights}


\end{frontmatter}

\section{Introduction}
\label{sec:introduction}

Prostate cancer is the second most commonly diagnosed cancer in men worldwide and a leading cause of cancer-related death~\cite{bray2024globocan}. The clinically decisive task is to separate \emph{clinically significant} prostate cancer (csPCa), which warrants treatment, from indolent disease: a missed aggressive lesion delays curative therapy, whereas over-calling benign or indolent tissue funnels patients toward unnecessary biopsies and the lasting harms of overtreatment, including urinary incontinence and erectile dysfunction. Because the diagnostic reference---systematic needle biopsy---is itself invasive and carries risks of hematuria, infection, and sepsis~\cite{loeb2013biopsy,bjurlin2013biopsy}, multi-parametric MRI (mpMRI), standardized by PI-RADS~\cite{turkbey2019pirads}, has become the recommended non-invasive first-line test, with landmark trials demonstrating that mpMRI-directed pathways reduce unnecessary biopsies while improving csPCa detection~\cite{ahmed2017promis,kasivisvanathan2018precision}. Expert mpMRI interpretation is, however, time-consuming and subject to non-trivial inter-reader variability, motivating automated csPCa segmentation as reproducible decision support.

Such automated systems must fuse the complementary mpMRI sequences---T2-weighted (T2W), apparent diffusion coefficient (ADC), and high $b$-value diffusion-weighted (HBV) imaging. A practical tension underlies this fusion: the diffusion-derived sequences (HBV, ADC) that carry the dominant csPCa signal under PI-RADS~\cite{turkbey2019pirads} are also the most fragile, being routinely degraded by patient motion, rectal-gas susceptibility artifacts, and scanner-specific variation~\cite{giganti2022piqual,plodeck2020rectalgas}, while abbreviated protocols or quality-control failures can render an acquisition non-diagnostic or altogether unavailable~\cite{hotker2022abbreviated,appayya2018ukconsensus}. The most diagnostically informative inputs are thus the least reliable at the point of care, and a model that silently mis-segments when one sequence is corrupted or missing is unsafe to deploy. Robustness to per-modality degradation is therefore a prerequisite---not an optional refinement---for clinical adoption.

Among fusion strategies, \emph{learned adaptive fusion}---in which a network weights or gates per-modality contributions instead of naively concatenating them---has become widely adopted for incomplete and multi-modal segmentation, with representative architectures including the transformer-based mmFormer~\cite{zhang2022mmformer}, the cross-modal attention CMAF-Net~\cite{sun2024cmaf}, and the region-aware RFNet~\cite{ding2021rfnet}. A premise common to the gated variants of this literature is that learned gates implement \emph{per-sample modality quality routing}: gate weights are assumed to dynamically reflect input-specific contrast quality or modality availability. This premise justifies the parameter overhead of gating modules and their addition atop otherwise-strong backbones. Yet the premise has rarely been tested directly: gate outputs are seldom inspected, and whether gate behavior depends on the inductive biases of the underlying backbone remains uncharacterized.

We address this gap through a systematic mechanism analysis of modality-isolated gated fusion (MIGF) for csPCa segmentation, organized along two pillars. First, a factorial ablation over eight module configurations---spanning gating (G), modality dropout (M), and deep supervision (D)---under 5-fold cross-validation across two architecturally distinct backbones, nnU-Net~\cite{isensee2021nnunet} (convolutional) and Mamba~\cite{gu2023mamba} (state-space), yielding 180 trained models on the PI-CAI dataset~\cite{saha2024picai}. Second, a direct gate mechanism analysis on 30 trained models, quantifying the per-sample variability of the learned gate weights to distinguish static modality priors from sample-dependent routing, and counterfactually replacing learned gates with training-set means. We further assess generalization under domain shift via cross-cohort evaluation on Prostate158~\cite{adams2022prostate158}. Contemporaneous prostate-MRI segmentation works advance methodologically orthogonal axes---spatiotemporal state-space modeling~\cite{zhao2026pcamamba}, multi-center pretraining scale~\cite{alzategrisales2026bimcv}, and multi-task PI-RADS grading~\cite{li2026frontiers}---none of which examine the backbone-conditional behavior of fusion gates that is our focus.

Our contributions are threefold:
\begin{enumerate}
    \item A systematic dual-backbone factorial ablation of gated fusion components under 5-fold cross-validation (180 models), quantifying each component's marginal effect and its consistency across architectural classes.
    \item Direct empirical evidence that learned gates are backbone-conditional: gate variance and counterfactual analyses show that convolutional backbones collapse gates into static modality priors while state-space backbones retain sample-dependent routing, challenging the prevailing routing assumption.
    \item Identification of architecture-dependent robustness behavior, including the cross-cohort specificity retention of MIGF-Mamba among the tested configurations and the benefit of training-time modality dropout across both tested backbones, yielding actionable guidance for clinical multi-modal MRI segmentation.
\end{enumerate}

\section{Methods}
\label{sec:methods}

\subsection{Datasets and preprocessing}
\label{sec:methods:data}
We used the PI-CAI dataset~\cite{saha2024picai} ($n=1500$ biparametric MRI studies, 404 with clinically significant prostate cancer), with positive-case lesion annotations combining the original human-expert delineations with the 205 manually annotated AI-positive cases reported by Pooch et al.~\cite{pooch2026semisupervised}, partitioned into five patient-level folds using the \texttt{picai\_baseline} pipeline (per-fold validation positives: 84, 72, 85, 77, 86). Each study comprises three sequences---T2W and HBV (Z-score normalized) and ADC (min--max scaled to $[0,1]$)---resampled and center-cropped to $128\times128\times32$ voxels. For cross-cohort evaluation we used Prostate158~\cite{adams2022prostate158} ($n=158$, 102 positive), processed identically; Prostate158 was used only for inference, never for training. Voxel- and case-level annotations follow their respective annotation protocols.

\subsection{Architecture and module configurations}
\label{sec:methods:arch}
\begin{figure*}[!t]
\centering
\includegraphics[width=\textwidth]{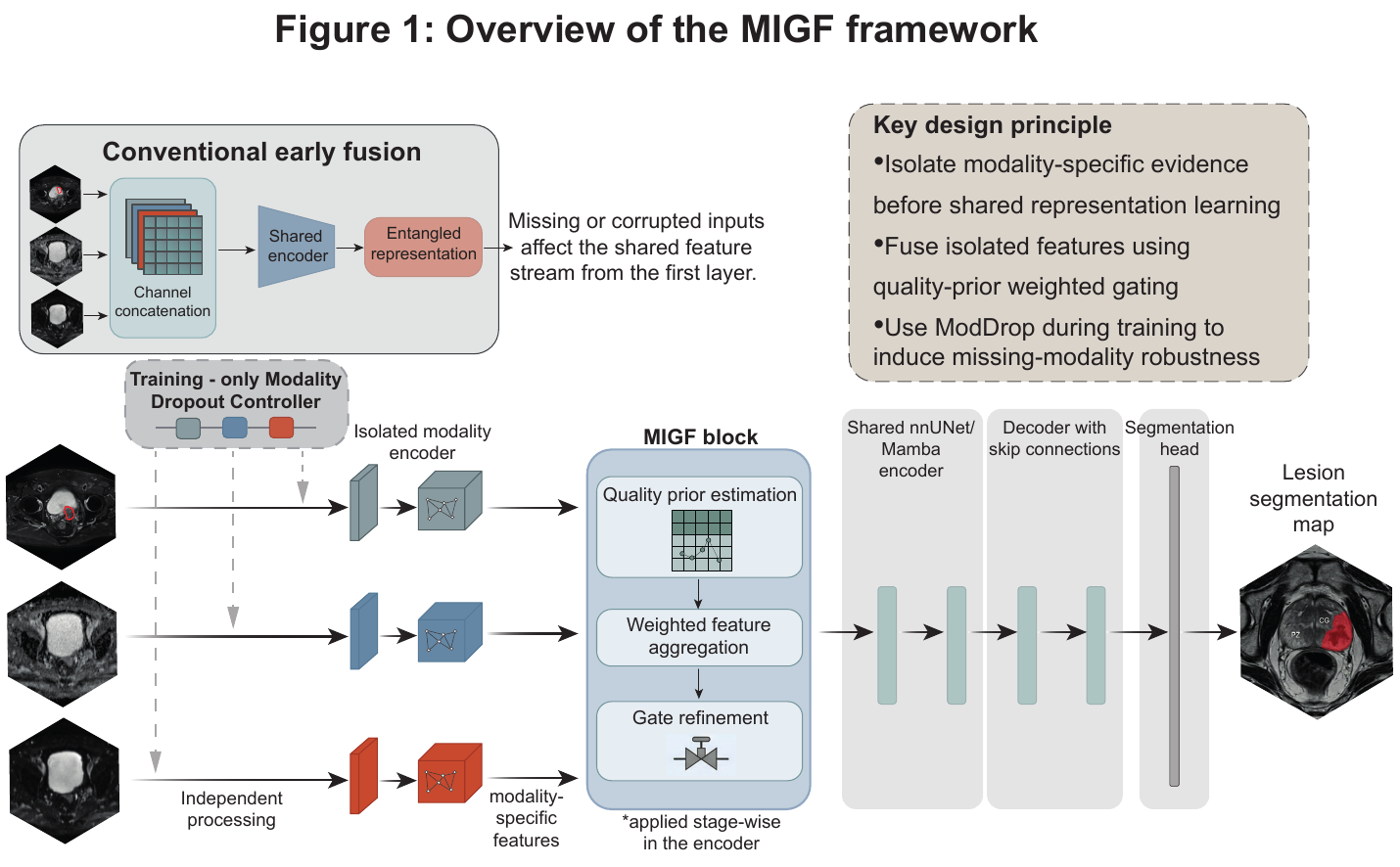}
\caption{Overview of the Modality-Isolated Gated Fusion (MIGF) framework. The
inset illustrates conventional early fusion, where T2W, HBV, and ADC inputs are
concatenated before a shared encoder, allowing a missing or corrupted modality
to affect the shared feature stream from the first layer. MIGF instead processes
each modality through an isolated, bias-free encoder and fuses the resulting
modality-specific features through the gating module; a zero-filled missing
modality produces identically zero features. During training, modality dropout
randomly zeroes one input modality; this is bypassed at inference. The fused
representation is passed to a shared backbone (nnU-Net or Mamba) and
segmentation head.}
\label{fig:overview}
\end{figure*}

Modality-Isolated Gated Fusion (MIGF) decouples per-modality feature extraction from fusion (Fig.~\ref{fig:overview}). Each modality ($m\in\{\text{T2W},\text{HBV},\text{ADC}\}$) is encoded by an independent, bias-free 3D-convolutional stream. Because these streams use no additive bias and only zero-preserving normalization and activations, a zero-filled (missing) modality yields an identically zero feature map, so an absent stream contributes neither signal nor noise to the fusion---the \emph{isolation property} that, by construction, lets the network degrade gracefully rather than catastrophically when an input is unavailable.

\begin{figure*}[!t]
\centering
\includegraphics[width=\textwidth]{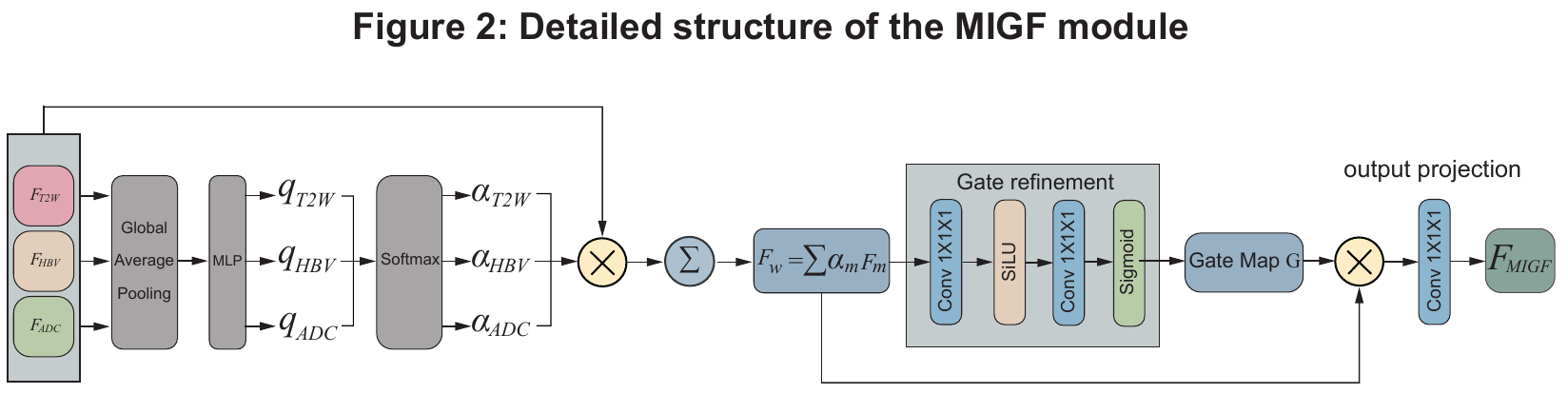}
\caption{Detailed structure of the MIGF module.
Modality-specific feature maps are reduced by global average pooling and
lightweight per-modality estimators to one score each, normalized by a softmax
into convex modality weights $\boldsymbol{\alpha}$ that linearly combine the
streams. A small feature-level gate---two $1\times1\times1$ convolutions with a
SiLU and a sigmoid---then rescales the combined feature element-wise, followed by
an output projection. The gate-mechanism analysis (Section~\ref{sec:methods:gate})
targets the modality weights $\boldsymbol{\alpha}$.}
\label{fig:migf_block}
\end{figure*}

The fusion module (\texttt{AdaptiveModalGating}, Fig.~\ref{fig:migf_block}) combines the per-modality features in two stages. First, a lightweight per-modality estimator scores each stream and a softmax turns the three scores into convex \emph{modality weights} $\boldsymbol{\alpha}$ (one non-negative weight per modality, summing to one) that linearly combine the streams; these weights, and whether they vary from case to case, are the object of our gate-mechanism analysis (Section~\ref{sec:methods:gate}). Second, a small feature-level gate---two $1\times1\times1$ convolutions with an intermediate SiLU and a sigmoid output---rescales the combined feature element-wise before a final projection. The module is lightweight, adding $2.34$M parameters on nnU-Net ($7.11\!\rightarrow\!9.45$M). Modality dropout (ModDrop)~\cite{neverova2016moddrop} stochastically zeroes one modality per sample with probability $0.3$ during training, exercising the isolation property. Deep supervision (D), when enabled, attaches two auxiliary $1\times1\times1$-convolution segmentation heads at the two intermediate decoder levels, whose losses are added to the main-output loss with weights $0.5$ and $0.25$.

We instantiate MIGF on two architecturally distinct backbones to contrast classes that differ in their implicit treatment of input channels: nnU-Net~\cite{isensee2021nnunet} (convolutional) and a Mamba state-space backbone~\cite{gu2023mamba}. Three components---gating (G), modality dropout (M), and deep supervision (D)---are studied in a full $2^3=8$ factorial: Bare (none), A1 (D+M), A2 (G+M), A3 (G+D), A4 (G), A5 (D), A6 (M), and Full (G+D+M) (Table~\ref{tab:ablation}). To test whether the cross-cohort behavior is specific to nnU-Net or shared across convolutional backbones, we additionally trained a third backbone---a MONAI 3D U-Net~\cite{cardoso2022monai,cicek20163dunet}---in its Bare and MIGF (A2) forms over the same folds and seeds; it enters the cross-cohort and efficiency comparisons (Tables~\ref{tab:crosscohort} and~\ref{tab:efficiency}) but not the factorial ablation.

\subsection{Training protocol}
\label{sec:methods:train}
All models were trained for 300 epochs (no early stopping) following the \texttt{picai\_baseline} recipe~\cite{saha2024picai}, using the AdamW optimizer~\cite{loshchilov2019adamw} (learning rate $5\times10^{-5}$) and a DiceFocal loss ($\alpha=0.9$, $\gamma=2.0$) combining region overlap with a focal term for class imbalance~\cite{milletari2016vnet,lin2017focal}, with batch size 8, automatic mixed precision, and deterministic cuDNN; checkpoints were selected by best validation Ranking Score. The 180 training runs were parallelized by dispatching one single-GPU job per model across four NVIDIA RTX~5090 GPUs (up to four concurrently). For nnU-Net, all eight configurations were trained with three seeds (42/123/789). For Mamba, the baseline and the primary MIGF configuration (A2) used three seeds, while the remaining six configurations used a single seed, reflecting compute budget (Mamba training is $7.6\times$ slower per epoch than nnU-Net) and following the single-seed convention common in recent prostate-MRI segmentation studies~\cite{zhao2026pcamamba,alzategrisales2026bimcv,li2026frontiers}. This yielded 180 trained models.

\subsection{Gate mechanism analysis}
\label{sec:methods:gate}
To test whether learned gates implement per-sample modality quality routing, we analyzed the modality weights $\boldsymbol{\alpha}$ introduced in Section~\ref{sec:methods:arch}---the component that embodies the per-sample modality routing the gating is assumed to perform---in the 30 trained A2 models (15 nnU-Net, 15 Mamba). For each model we extracted the entry-stage weights $\alpha_m$ on the validation set and quantified per-sample variability by the \emph{across-case standard deviation} $\sigma_{\text{case}}(\alpha_m)$ (averaged over modalities): a small value indicates a sample-invariant (static) prior, a large value sample-dependent routing. As a complementary functional test, we performed a \emph{counterfactual replacement}: at inference the per-sample gate output was replaced by the training-set mean weight $\bar{\boldsymbol{\alpha}}$. We define $\Delta$ as the metric obtained with the learned per-sample gate minus that obtained after mean-gate replacement, so a positive $\Delta$ indicates that the per-sample gate outperforms its static mean (equivalently, that the replacement degrades performance).

\subsection{Evaluation and statistics}
\label{sec:methods:stats}
Detection was scored with \texttt{picai\_eval}~\cite{picai_eval2024}, reporting the Ranking Score (the mean of lesion-level AUROC and average precision), together with case-level specificity (CaseSpec), case-level sensitivity (CaseSens), and voxel-level Dice. For each configuration we computed paired differences against the Bare baseline over matched (fold, seed) pairs and tested them with a two-sided paired $t$-test and the Wilcoxon signed-rank test, reporting 95\% confidence intervals on the mean difference; for the six single-seed Mamba configurations ($n=5$) we report descriptive statistics and note that the Wilcoxon test is underpowered (minimum attainable two-sided $p=0.0625$). The Bare-versus-A2 contrast was the pre-specified primary comparison; the remaining configuration contrasts are exploratory, and $p$-values are not adjusted for multiple comparisons. Marginal component effects were computed as the mean Ranking Score of the four configurations containing a component minus that of the four without it. All summary statistics use the unbiased standard deviation (ddof~$=1$).

\section{Results}
\label{sec:results}

\subsection{Gating is backbone-conditional}
\label{sec:results:ablation}
Table~\ref{tab:ablation} reports the eight-configuration ablation on both backbones (ideal-input scenario). On nnU-Net, adding gating consistently \emph{reduced} the Ranking Score: the three gating-containing configurations that reached significance---A2 (G+M), A3 (G+D), and A4 (G)---scored $-0.036$ ($p=0.012$), $-0.060$ ($p=0.003$), and $-0.054$ ($p=0.006$) versus Bare (two-sided paired $t$-test; Wilcoxon $p=0.016$/$0.008$/$0.010$). The best nnU-Net configuration was A1 (D+M, no gating) at $0.7483\pm0.043$, exceeding Bare ($0.7363\pm0.041$). The marginal component effects confirmed this pattern: gating $-0.037$, deep supervision $-0.003$, modality dropout $+0.023$.

On Mamba the direction reversed for gating. The primary MIGF configuration A2 (G+M) significantly improved the Ranking Score ($+0.024$, $p=0.037$; Wilcoxon $p=0.044$) and yielded the largest case-level specificity of any Mamba configuration ($0.579$ vs.\ Bare $0.465$); this $+0.114$ specificity gain was directionally consistent but did not reach significance (95\% CI $[-0.009,+0.237]$, $p=0.067$). The Mamba marginal effects were gating $+0.005$, deep supervision $-0.013$, and modality dropout $+0.021$. Modality dropout was thus the only component with a positive marginal effect on \emph{both} tested backbones ($+0.023$ and $+0.021$). The six single-seed Mamba configurations ($n=5$) are reported descriptively (Table~\ref{tab:ablation}) and were not significance-tested.

\subsection{Gate mechanism: static prior versus per-sample routing}
\label{sec:results:gate}
\begin{figure}[!t]
\centering
\includegraphics[width=\columnwidth]{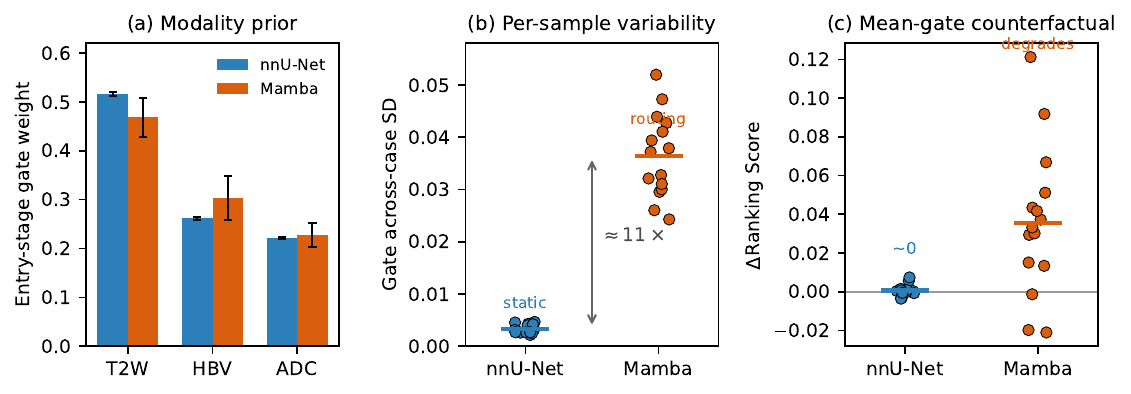}
\caption{Gate-mechanism analysis on the 30 A2 models. (a) Both backbones learn a
similar entry-stage modality prior (mean gate weights for T2W/HBV/ADC; error
bars are the across-case SD). (b) The across-case SD of the gate weights---their
per-sample variability---separates the two backbones by $\approx$11$\times$ with
non-overlapping distributions: nnU-Net gates are essentially static
($0.003$), Mamba gates vary per sample ($0.037$). (c) Replacing each per-sample
gate output with the training-set mean leaves nnU-Net unchanged ($\Delta$Ranking
Score $\approx0$, static prior) but degrades Mamba (per-sample routing).
Horizontal bars: means; points: individual checkpoints ($n=15$ per backbone).}
\label{fig:gatescatter}
\end{figure}

The gate-weight analysis (30 A2 models; Table~\ref{tab:gate}, Fig.~\ref{fig:gatescatter}) revealed a sharp architectural dichotomy. Both backbones converged to a similar entry-stage modality prior (nnU-Net weights T2W/HBV/ADC $0.52$/$0.26$/$0.22$; Mamba $0.47$/$0.30$/$0.23$), but their per-sample variability differed by roughly an order of magnitude. nnU-Net gate weights had an across-case standard deviation of $0.0033\pm0.0008$, essentially invariant across samples---a static modality prior. Mamba gate weights varied $\approx$11$\times$ more across samples ($0.0365\pm0.0080$), indicating sample-dependent routing; the two distributions did not overlap (nnU-Net maximum $0.0047$ vs.\ Mamba minimum $0.0243$).

The counterfactual replacement corroborated this. Replacing each model's per-sample gate output with its training-set mean left nnU-Net essentially unchanged ($\Delta$Ranking Score $+0.001\pm0.003$, with $|\Delta|<0.01$ for all 15 models; $\Delta$Dice $\approx 0$), confirming that the learned per-sample gating carries negligible information on nnU-Net. On Mamba the same replacement degraded performance ($\Delta$Ranking Score $+0.036\pm0.038$, $\Delta$CaseSpec $+0.084$, $\Delta$Dice $+0.043$ positive for all 15 models), confirming that Mamba gates encode useful per-sample routing. Together these results show that learned gating implements per-sample quality routing only on the state-space backbone; on the convolutional backbone it collapses into a static prior.

\subsection{Cross-cohort generalization}
\label{sec:results:crosscohort}
Evaluating the primary models on Prostate158 without retraining (Table~\ref{tab:crosscohort}) exposed an architecture-conditional failure. Both convolutional backbones collapsed to case-level specificity near zero---nnU-Net (bare $0.000$, MIGF $0.004$) and a MONAI 3D U-Net (bare $0.002$, MIGF $0.011$)---despite moderate cross-cohort Ranking Scores ($0.39$--$0.47$). In contrast, both Mamba families retained substantial specificity (bare $0.280$, MIGF-Mamba $0.314$), with MIGF-Mamba having the highest case-level specificity. MIGF also improved the cross-cohort Ranking Score over the corresponding bare backbone for all three backbones (nnU-Net $+0.047$, U-Net $+0.029$, Mamba $+0.101$), indicating that modality isolation aids out-of-distribution transfer even where overall specificity collapses.

\subsection{Computational cost}
\label{sec:results:efficiency}
The MIGF module is lightweight in parameters but its cost is dominated by the backbone (Table~\ref{tab:efficiency}; single case, batch~1, RTX~5090). MIGF-nnU-Net adds modest overhead (9.45M parameters, $8.5$~ms inference) relative to bare nnU-Net ($7.11$M, $4.2$~ms), whereas MIGF-Mamba is an order of magnitude slower ($18.7$M, $107$~ms) owing to the sequential state-space scan. This $12.5\times$ inference gap between the two MIGF variants is relevant to deployment (Section~\ref{sec:discussion}).


\begin{figure}[!t]
\centering
\includegraphics[width=\columnwidth]{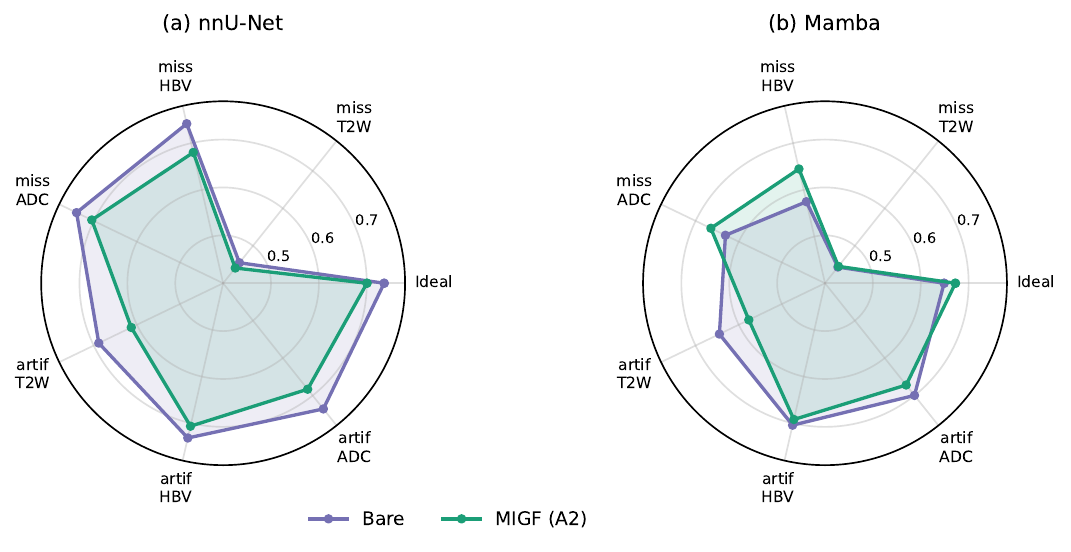}
\caption{Seven-scenario robustness (5-fold mean Ranking Score), Bare versus
MIGF (A2), for (a) nnU-Net and (b) Mamba. Adding gating reduces nnU-Net's score
in every scenario, whereas on Mamba it benefits the missing-diffusion scenarios
(missing-HBV, missing-ADC), illustrating that the backbone-conditional gating
effect persists under modality degradation.}
\label{fig:robustness}
\end{figure}


\begin{table}[!t]
\centering
\caption{Eight-configuration factorial ablation on the PI-CAI ideal-input scenario,
reported as mean\,$\pm$\,SD (ddof\,$=1$) of the Ranking Score and case-level
specificity (CaseSpec). Components: G\,=\,gating, D\,=\,deep supervision,
M\,=\,modality dropout. nnU-Net configurations use 3 seeds $\times$ 5 folds
($n=15$); Mamba Bare and A2 use 3 seeds ($n=15$), the remaining six Mamba
configurations use a single seed ($n=5$). $p$-values are two-sided paired
$t$-tests of the Ranking Score difference versus Bare (n.s.\ for $n=5$
configurations, which are underpowered).}
\label{tab:ablation}
\small
\begin{tabular}{lcccccc}
\toprule
 & G & D & M & \multicolumn{2}{c}{Ranking Score} & CaseSpec \\
\cmidrule(lr){5-6}
Config & & & & mean\,$\pm$\,SD & $p$ vs Bare & mean\,$\pm$\,SD \\
\midrule
\multicolumn{7}{l}{\emph{nnU-Net} ($n=15$ per config)} \\
Bare & -- & -- & -- & $0.7363\pm0.041$ & --     & $0.534\pm0.198$ \\
A1   & -- & Y  & Y  & $0.7483\pm0.043$ & $0.44$ & $0.535\pm0.164$ \\
A2   & Y  & -- & Y  & $0.7004\pm0.048$ & $0.012$ & $0.456\pm0.153$ \\
A3   & Y  & Y  & -- & $0.6764\pm0.067$ & $0.003$ & $0.445\pm0.197$ \\
A4   & Y  & -- & -- & $0.6826\pm0.054$ & $0.006$ & $0.401\pm0.185$ \\
A5   & -- & Y  & -- & $0.7016\pm0.053$ & $0.058$ & $0.515\pm0.145$ \\
A6   & -- & -- & Y  & $0.7294\pm0.046$ & $0.71$ & $0.426\pm0.225$ \\
Full & Y  & Y  & Y  & $0.7095\pm0.052$ & $0.097$ & $0.453\pm0.149$ \\
\midrule
\multicolumn{7}{l}{\emph{Mamba} (Bare, A2: $n=15$; others: $n=5$)} \\
Bare & -- & -- & -- & $0.6488\pm0.050$ & --     & $0.465\pm0.181$ \\
A1   & -- & Y  & Y  & $0.6385\pm0.065$ & n.s.   & $0.341\pm0.152$ \\
A2   & Y  & -- & Y  & $0.6724\pm0.055$ & $0.037$ & $0.579\pm0.091$ \\
A3   & Y  & Y  & -- & $0.6544\pm0.060$ & n.s.   & $0.534\pm0.148$ \\
A4   & Y  & -- & -- & $0.6217\pm0.069$ & n.s.   & $0.503\pm0.162$ \\
A5   & -- & Y  & -- & $0.6208\pm0.081$ & n.s.   & $0.360\pm0.244$ \\
A6   & -- & -- & Y  & $0.6708\pm0.054$ & n.s.   & $0.540\pm0.100$ \\
Full & Y  & Y  & Y  & $0.6487\pm0.066$ & n.s.   & $0.514\pm0.302$ \\
\bottomrule
\end{tabular}
\end{table}

\begin{table}[!t]
\centering
\caption{Gate mechanism analysis on the 30 A2 models (15 per backbone).
The across-case standard deviation of the entry-stage modality weights
quantifies per-sample variability (small = static prior, large = per-sample
routing). Counterfactual replacement substitutes each per-sample gate output
with the training-set mean; $\Delta$ is the learned per-sample gate's
performance minus that after replacement, so a positive $\Delta$ means the
per-sample gate helps (i.e., replacement degrades performance). All
values mean\,$\pm$\,SD over 15 models.}
\label{tab:gate}
\small
\begin{tabular}{lcc}
\toprule
 & nnU-Net & Mamba \\
\midrule
Gate across-case SD                & $0.0033\pm0.0008$ & $0.0365\pm0.0080$ \\
\quad range                        & $[0.0022,0.0047]$ & $[0.0243,0.0520]$ \\
$\Delta$Ranking Score (mean-gate)  & $+0.001\pm0.003$ & $+0.036\pm0.038$ \\
$\Delta$CaseSpec (mean-gate)       & $-0.001\pm0.003$ & $+0.084\pm0.056$ \\
$\Delta$Dice (mean-gate)           & $-0.000\pm0.000$ & $+0.043\pm0.019$ \\
\quad models with $|\Delta\text{Score}|<0.01$ & 15/15 & 1/15 \\
\bottomrule
\end{tabular}
\end{table}

\begin{table}[!t]
\centering
\caption{Cross-cohort evaluation on Prostate158 ($n=158$, 102 positive)
without retraining, mean over 5 folds $\times$ 3 seeds ($n=15$ per family).
PI-CAI ideal-scenario Ranking Score is shown for reference.}
\label{tab:crosscohort}
\small
\begin{tabular}{lccc}
\toprule
Family & PI-CAI Score & Prostate158 Score & Prostate158 CaseSpec \\
\midrule
Bare nnU-Net    & $0.736$ & $0.394$ & $0.000$ \\
MIGF-nnU-Net A2 & $0.700$ & $0.441$ & $0.004$ \\
Bare UNet       & $0.736$ & $0.437$ & $0.002$ \\
MIGF-UNet A2    & $0.719$ & $0.466$ & $0.011$ \\
Bare Mamba      & $0.649$ & $0.345$ & $0.280$ \\
MIGF-Mamba A2   & $0.672$ & $0.446$ & $0.314$ \\
\bottomrule
\end{tabular}
\end{table}

\begin{table}[!t]
\centering
\caption{Computational cost (single case, batch~1, RTX~5090, fp16 autocast;
mean\,$\pm$\,SD over 10 timed passes after 3 warmup passes). Latency is
dominated by the backbone class.}
\label{tab:efficiency}
\small
\begin{tabular}{lccc}
\toprule
Model & Params (M) & Latency (ms) & Peak mem.\ (MB) \\
\midrule
Bare nnU-Net    & $7.11$  & $4.2\pm0.1$   & $346$ \\
MIGF-nnU-Net A2 & $9.45$  & $8.5\pm0.1$   & $523$ \\
Bare UNet       & $31.80$ & $2.4\pm0.0$   & $234$ \\
MIGF-UNet A2    & $52.60$ & $12.3\pm0.1$  & $702$ \\
Bare Mamba      & $9.82$  & $52.0\pm0.5$  & $293$ \\
MIGF-Mamba A2   & $18.67$ & $106.5\pm0.9$ & $348$ \\
\bottomrule
\end{tabular}
\end{table}

\section{Discussion}
\label{sec:discussion}

\subsection{Why gating behavior is backbone-conditional}
Our central finding is that the same gated-fusion module behaves oppositely on two backbones, and the gate-mechanism analysis offers a mechanistic explanation. On nnU-Net, the learned gate degenerated into a static modality prior (across-case gate-weight SD $0.0033$; counterfactual replacement with the training-set mean left performance unchanged, $\Delta$Score $+0.001$), and adding gating reduced the Ranking Score (marginal effect $-0.037$; A2/A3/A4 all $p<0.05$). We attribute this to the convolutional backbone already encoding modality-specific weighting implicitly: 3D convolutions process the three sequences as input channels with per-channel filters and per-channel normalization, so an explicit gating MLP is largely redundant and its $2.34$M additional parameters interfere with, rather than aid, optimization. On Mamba the gate retained sample-dependent variation (across-case SD $0.0365$, $\approx$11$\times$ higher; mean-gate replacement degraded performance, $\Delta$Dice $+0.043$ in 15/15 models) and the G+M configuration significantly improved the Ranking Score ($+0.024$, $p=0.037$). State-space kernels, designed for sequence modeling, lack a built-in mechanism for differential channel weighting, so the explicit gate supplies a missing inductive bias. The $\approx$11$\times$ separation in per-sample gate variability between the two backbones, with non-overlapping distributions, indicates that whether a learned gate performs per-sample routing or collapses to a static prior is determined by the inductive bias of the underlying architecture---not by the gating module in isolation. This reframes a common assumption in gated multi-modal fusion: the routing interpretation is architecture-contingent rather than universal.

\subsection{Clinical and deployment implications}
Two findings bear directly on deployment. First, modality dropout was the only component with a positive marginal effect on both tested backbones ($+0.023$, $+0.021$), supporting training-time stochastic modality removal as an architecture-agnostic robustness measure for protocols with variable diffusion acquisitions. The backbone-conditional pattern also persisted under simulated modality degradation (Fig.~\ref{fig:robustness}): across the seven evaluation scenarios, adding gating reduced nnU-Net's Ranking Score in every scenario, whereas on Mamba it specifically benefited the clinically relevant missing-diffusion cases (missing-HBV $+0.070$, missing-ADC $+0.034$), consistent with the routing role of its gates. Second, cross-cohort behavior was strongly architecture-dependent: on Prostate158, both Mamba variants retained case-level specificity ($0.280$, $0.314$) whereas nnU-Net collapsed to near zero. This collapse was not specific to nnU-Net---a second convolutional backbone (a MONAI 3D U-Net~\cite{cardoso2022monai,cicek20163dunet}) showed the same near-zero cross-cohort specificity (Table~\ref{tab:crosscohort}), indicating the failure is tied to the convolutional architecture class rather than a single model. Clinically this is consequential: a backbone that loses case-level specificity on data from an unseen scanner would flag a large share of benign studies as suspicious at a new site, generating precisely the unnecessary biopsies that mpMRI triage exists to prevent---so retained cross-cohort specificity is a safety-relevant property, not a cosmetic metric. MIGF-Mamba was therefore the only tested configuration combining competitive in-distribution performance with retained out-of-distribution specificity, and modality isolation further improved cross-cohort Ranking Score for every backbone. However, this robustness carries a cost: MIGF-Mamba inference is $12.5\times$ slower than MIGF-nnU-Net ($107$ vs.\ $8.5$~ms; Table~\ref{tab:efficiency}). Backbone selection should thus balance cross-cohort robustness against the inference budget of the target clinical workflow.

\subsection{Relation to prior work}
Recent prostate-MRI segmentation studies advance orthogonal directions: spatiotemporal state-space modeling~\cite{zhao2026pcamamba}, multi-center pretraining at scale~\cite{alzategrisales2026bimcv}, and joint segmentation with PI-RADS grading~\cite{li2026frontiers}. None examine the mechanism of fusion gates or report backbone-conditional behavior; our contribution is complementary, informing the fusion-module design space that such systems inherit. Relative to adaptive multi-modal fusion architectures such as mmFormer~\cite{zhang2022mmformer} and RFNet~\cite{ding2021rfnet}, our analysis suggests that reported gains may depend on the backbone's implicit modality handling, motivating gate-output inspection as a routine diagnostic for the gated members of this family.

\subsection{Limitations}
Several limitations qualify these findings. First, case-level specificity differences, although directionally consistent, were generally not individually significant owing to high inter-fold variance (e.g., the MIGF-Mamba specificity gain of $+0.114$ had $p=0.067$); the robust claims rest on the Ranking Score and the gate-mechanism analysis. Second, owing to Mamba's $7.6\times$ higher training cost, six of its eight configurations used a single seed and are reported descriptively; the primary Bare-versus-A2 comparison and the gate analysis, however, used three seeds on both backbones. Third, the gate-mechanism analysis was performed on the A2 configuration only and on a per-stage scalar gate; per-voxel or attention-style gates may behave differently. Fourth, cross-cohort evaluation used a single external dataset (Prostate158); broader multi-center validation is warranted. Finally, our conclusions concern biparametric MRI; extension to dynamic-contrast or other modality sets remains future work.

\section{Conclusion}
\label{sec:conclusion}

Through a factorial ablation and a direct gate-mechanism analysis across two architecturally distinct backbones, we showed that learned modality gating in multi-modal prostate MRI segmentation is backbone-conditional rather than universal. On the convolutional nnU-Net, gates collapsed into a static modality prior and degraded performance, whereas on the state-space Mamba they retained sample-dependent routing and improved it---an $\approx$11$\times$ separation in per-sample gate-weight variability with non-overlapping distributions. The routing interpretation that motivates much gated-fusion work is therefore contingent on the backbone's implicit modality handling, and gate-output inspection is a worthwhile routine diagnostic. Of the three studied components, only training-time modality dropout was beneficial across both tested backbones, providing an architecture-agnostic robustness measure. Finally, MIGF-Mamba was the only tested configuration combining competitive in-distribution accuracy with retained case-level specificity under cross-cohort shift, at the cost of an order-of-magnitude higher inference latency; backbone selection for clinical deployment should weigh this robustness against the available compute budget.

\par


\section*{CRediT authorship contribution statement}
\textbf{Yongbo Shu:} Conceptualization, Methodology, Formal analysis, Validation, Visualization, Writing -- original draft, Writing -- review \& editing. \textbf{Wenzhao Xie:} Conceptualization. \textbf{Shanhu Yao:} Methodology. \textbf{Zirui Xin:} Methodology. \textbf{Luo Lei:} Validation. \textbf{Kewen Chen:} Visualization. \textbf{Aijing Luo:} Supervision, Project administration, Funding acquisition, Writing -- review \& editing.

\section*{Declaration of competing interest}
The authors declare that they have no known competing financial interests or personal relationships that could have appeared to influence the work reported in this paper.

\section*{Data and code availability}
The source code for MIGFNet is publicly available at \url{https://github.com/yosh3289/MIGFNet}. All internal experiments were conducted on the publicly available PI-CAI dataset (\url{https://zenodo.org/records/6624726}). External validation used the Prostate158 dataset (\url{https://github.com/kbressem/prostate158}).

\section*{Acknowledgements}
This study was supported by the Graduate Innovation Program of Central South University and funded by Central South University (Project No.\ 1053320214354, Yongbo Shu). It was further supported by the Key Laboratory of Medical Information Research of Central South University within the project ``Clinical Research Center for Cardiovascular Intelligent Healthcare in Hunan Province'' (agreement no.\ 2021SK4005) and the Science and Technology Plan Project of Changsha (grant no.\ kq1901133). The authors thank Zihong Shu and Shuying Zhang for providing the computing resources used in this study.

\section*{Declaration of generative AI and AI-assisted technologies in the writing process}
During the preparation of this work, the authors used Claude (Anthropic) to assist with language refinement, grammar checking, and manuscript formatting. The authors manually reviewed all experimental procedures, verified the accuracy of all reported data, and confirmed the correctness of all cited references. After using this tool, the authors reviewed and edited the content as needed and take full responsibility for the content of the published article.

\bibliographystyle{elsarticle-num}
\bibliography{references}

\end{document}